\documentclass[a4paper, twocolumn, 10pt]{IEEEtran} 

\usepackage{graphicx}
\usepackage{mathrsfs}
\makeatletter
\let\NAT@parse\undefined

\usepackage{amsmath,amsfonts,amssymb}
\usepackage{url}
\usepackage{graphicx}
\usepackage{float}
\usepackage{multirow}
\usepackage{hyperref}
\usepackage{tikz}
\usepackage{pgf}
\usepackage{placeins}
\usepackage{subcaption}
\usepackage{comment}
\usepackage{url}
\usepackage{placeins}
\usepackage{latexsym}
\usepackage{pifont}
\usepackage{pgfplots,filecontents}
\usepackage{gensymb}
\pgfplotsset{compat=newest}

\usepackage[symbol]{footmisc}

\usetikzlibrary{intersections}

\newcommand{\R}{\mathbb{R}}

\newcommand{\E}{\mathbb{E}}

\DeclareGraphicsExtensions{.png,.jpg,.jpeg,.eps,.gif}
\usepackage{cleveref}

\DeclareGraphicsExtensions{.pdf,.png,.jpg,.jpeg,.eps}
\title{Memory-based Deep Reinforcement Learning for Obstacle Avoidance in UAV with Limited Environment Knowledge}

\author{Abhik Singla*, Sindhu Padakandla and Shalabh Bhatnagar
\thanks{We thank NVIDIA Corporation for the TitanX GPU used for this research.}
\thanks{The authors are with the Department of Computer Science and Automation and Robert Bosch Centre for Cyber-Physical Systems, Indian Institute of Science, Bangalore, India. 
        {E-mail: \tt\small \{abhiksingla,sindhupr,shalabh\}@iisc.ac.in}. *Corresponding author}%
}
\begin{document}
\nocite{*}
\maketitle

\thispagestyle{empty}
\pagestyle{empty}


\begin{abstract}
This paper presents our method for enabling a UAV
quadrotor, equipped with a monocular camera, to autonomously avoid collisions with obstacles in 
unstructured and unknown indoor
environments.  When compared to obstacle  avoidance  in  ground  vehicular robots, 
UAV navigation brings in additional challenges because the UAV
motion is no more constrained to a well-defined indoor ground or
street environment. Horizontal structures in indoor and outdoor
environments like decorative items, furnishings, ceiling fans, sign-
boards, tree branches etc., also become relevant obstacles unlike
those  for  ground  vehicular  robots.  Thus,  methods  of  obstacle
avoidance developed for ground robots are clearly inadequate for
UAV  navigation.  Current control  methods  using  monocular  images  for
UAV  obstacle  avoidance  are  heavily  dependent  on  environment
information.  
These controllers do not fully retain and utilize the extensively available information about the ambient environment for
decision making. We propose a deep reinforcement learning based
method for UAV obstacle avoidance (OA) and autonomous exploration
which  is  capable  of  doing  exactly  the  same.  The  crucial  idea  in
our method is the concept of partial observability and how UAVs
can retain relevant information about the environment structure
to  make better future navigation  decisions.  
Our  OA technique  uses recurrent neural networks with temporal attention and 
provides better results compared to prior works in terms of distance covered
during  navigation  without  collisions.  In  addition,  our  technique
has  a  high  inference  rate  (a  key  factor  in  robotic  applications)
and is energy-efficient as it minimizes oscillatory motion of UAV
and  reduces  power  wastage.
\end{abstract}

\section*{Supplementary Material}
For supplementary video see: \url{https://bit.ly/2PNgWsk}. The project's code is available at \url{https://github.com/abhiksingla/UAV_obstacle_avoidance_controller}
\section{Introduction}\label{sec:intro}
\begin{figure}
\begin{center}
\includegraphics[scale=0.17]{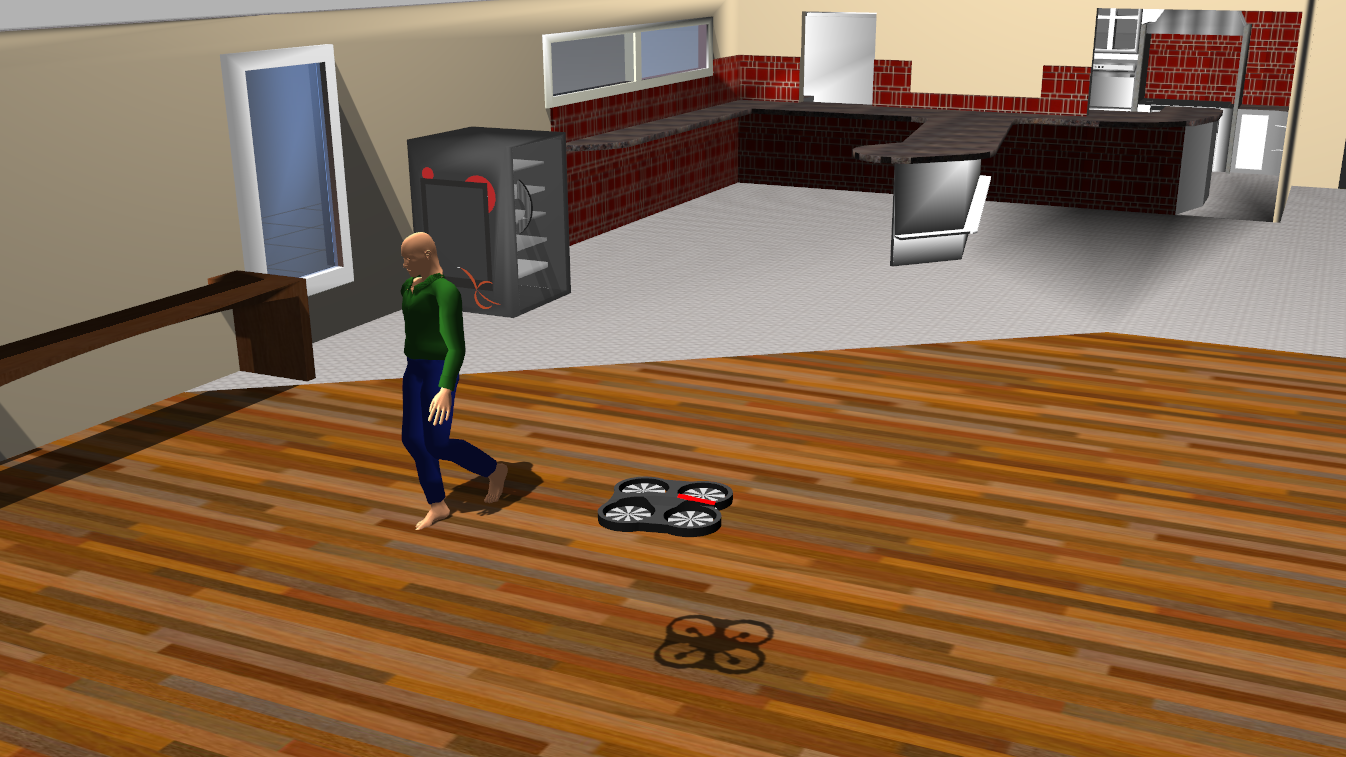}

\caption{A UAV encountering stationary as well as moving obstacles in an indoor environment. Here, the walking human being is a moving obstacle, whose direction and future intent of motion cannot be predicted.}
\label{fig:ss1}
\end{center}
\end{figure}

Unmanned aerial vehicles (UAVs) or ``drones" are cyber-physical systems that can be operated either by remote control (using a mobile application on a smartphone over a wireless channel) or autonomously using onboard computers. Ranging from crop \cite{crop} and infrastructure monitoring \cite{infra}, rescue operations and disaster management \cite{disaster}, to more popular uses like goods delivery and filming \cite{delivery,filming}, UAVs are increasingly finding their application in diverse scenarios. Owing to their small size and light weight, UAVs can penetrate into constricted spaces or effortlessly glide over pre-specified geographical areas, the majority of which may possibly be beyond the reach of humans. However, UAVs still lack some elementary capabilities which impede their widespread use. 
One such example is the ability to avoid obstacles. Avoiding obstacles is a non-trivial task because the obstacles might be so positioned that avoiding them requires delicate and dexterous movements.
To be able to avoid obstacles, the UAV must be able to perceive the distance between itself and the obstacles along with other visual cues such as the shape of the obstacle and it's height. This crucial visual information enables a UAV to infer traversable spaces and obstacles (see Fig. \ref{fig:ss1} for an illustration). 

Classical approaches for inferring visual geometry include techniques like Simultaneous Localization and Mapping (SLAM) and Structure from Motion (SfM). These techniques use measurements from sensors like Kinect \cite{kinect}, Light Detection and Ranging (LIDAR), Sound Navigation and Ranging (SONAR), optical flow, stereo and monocular cameras for computation. SLAM algorithms utilize measurements from a single sensor \cite{slam-rgbd} or a combination of sensors \cite{slam-lidar-stereo} to build or update a map of the environment surrounding the UAV while simultaneously using the same to estimate the UAV's position. The SfM approaches use measurements from sensors like optical flow \cite{sfm-opt-flow} and/or a moving monocular camera \cite{sfm-monocular} to determine depth map and the 3D structure.
SLAM and SfM approaches require the UAV to compute a path and then navigate through it. The UAV needs to repetitively hover, compute the depth map and then find a suitable path. Thus, path planning on the fly is not easy in SLAM and SfM approaches. This also means that SLAM and SfM approaches cannot be used for real-time obstacle avoidance based on the visual information gathered about the surroundings. \cite{realtimeslam} proposes a SLAM technique which computes a path on the fly. However, such an enhancement does not avoid dynamic and non-stationary obstacles whose movements cannot be predicted. Another disadvantage of using SLAM and SfM methods is that these do not detect untextured walls. Untextured walls normally arise in indoor environments and hence being able to distinguish textures on walls is crucial to obstacle avoidance.

Kinect, LIDAR, SONAR, optical flow and stereo camera sensors are widely used for depth estimation (see \cite{only-kinect,only-opt-flow}) and hence these can be potentially used for obstacle avoidance as well without resorting to computation-intensive approaches like SLAM and SfM. However these sophisticated sensors are expensive and add unnecessary burden to the UAV in terms of weight as well as consumption of power. Moreover, optical flow and stereo camera are not suited for long-range obstacle avoidance. Other sensors like for example, the monocular camera,  is essential for every UAV application, as it gives visual information.
The monocular camera is a low-cost sensor which provides RGB images of the UAV's ambient environment. In comparison to the heavy-weight sensors mentioned earlier, a monocular camera is light-weight. The question then is whether we can use a monocular camera for depth estimation as well and plausibly for obstacle avoidance. 


Extracting the range information (i.e., distance between the sensor and the various objects in front of the sensor) from the monocular RGB images is a challenging problem, simply because the camera captures only the 2-D information of the surrounding environment.
Some recent works \cite{fcrn}-\cite{cad2rl} address the issue of depth prediction using monocular camera RGB images by leveraging deep learning techniques. 
Supervised and semi-supervised learning approaches (\cite{fcrn}-\cite{egomotion}) collect huge amounts of data consisting of the monocular images and the corresponding depth maps to train a deep learning model. Such models are based on convolutional neural networks (CNN) or their variants (residual networks \cite{fcrn}). Given a single image, the deep network outputs the predicted depth map from the monocular image. The proposed approaches in \cite{fcrn}-\cite{egomotion} however do not tackle the vital problem of UAV obstacle avoidance and navigation, which is the problem that we are interested in this paper.

Varied obstacle avoidance techniques in conjunction with depth prediction are proposed in \cite{punarjay}-\cite{cad2rl}. \cite{punarjay} proposes a behavior arbitration scheme to 
obtain the yaw and pitch angles for the UAV to avoid an obstacle and for navigation in general.
Trajectory planning using obstacle bounding boxes and depth estimation is explored in \cite{jmod}.
This work designs a CNN architecture that jointly estimates depth and obstacle bounding boxes. The extracted information is then utilized in the RRT-Connect planner to plan trajectories between a start and end point. \cite{intermediate} proposes two different CNN architectures - one for depth and surface normal estimation and the other for trajectory prediction. Both the CNNs use a 3D cost function for training and evaluation. \cite{crash-learn} follows an unconventional approach, wherein the authors collect a dataset of UAV crashes. This dataset is labeled and then input to a CNN model. Given an image obtained from the monocular camera, the network predicts how the UAV should move in the next instant to avoid a crash. 
UAV navigation in the presence of obstacles is inherently a sequential decision making problem under uncertainty. This is because an action taken at an instant affects the path of the UAV in the future instants too. Hence, it is appropriate to design obstacle avoidance in UAV as a Reinforcement learning (RL) problem. CAD2RL \cite{cad2rl} proposes a Deep RL (DRL) method for obstacle avoidance in indoor flight. This work trains a UAV for navigation using simulated 3D hallway environments. For this, a large number of 3D hallway environment images with different lighting, wall textures, furniture placement are generated and a deep Q-network learns UAV movement policy on these images. However, this work requires substantial amount of data concerning the images of hallway environments and is not efficient. Moreover, the method proposed in \cite{cad2rl} is not intuitive. It does not attempt to mimic how humans learn to avoid obstacles. The basic information which helps the human brain to navigate is the depth information (owing to the binocular vision) and not the RGB information. 

Our work adds a new dimension to the existing work on UAV obstacle avoidance. We are motivated from how humans decide what to do next given a scenario. Humans have \emph{limited} or \emph{partial access} to the environment, but still are able to solve challenging problems in daily lives. All this is possible, because human brain has memory which is key to summarizing and storing relevant information for tackling problems. This memory is capable of effectively storing and recalling relevant information gathered over time in order to take the next suitable decision in every scenario. UAV obstacle avoidance and navigation also present a similar problem of \emph{partial observability} which requires a notion of memory.
For example, while navigating, a UAV may fly towards a corner. When it is approaching the corner, the depth map might indicate more space in the front when compared to the sides. The lack of temporal information coupled with limited field of vision of the monocular camera makes the UAV to move ahead towards the corner and crash onto the wall. Such scenarios are very common in UAV navigation and hence require a controller which can utilize the relevant past information.
Our aim is to design a UAV control algorithm which has the capability to combine information obtained over a period of time in order to make better navigation and obstacle avoidance decisions. 

We propose a deep RL method which enables the UAV controller to collect and store relevant observations gathered over time. This method is based on recurrent neural network (RNN) architecture with an additional function called \emph{Temporal Attention}. Using this architecture the UAV controller learns a control policy to avoid obstacles.

\subsection{Organization of the Paper}
The next section describes the method we have developed for UAV obstacle avoidance. Section \ref{sec:exp} gives the details of experimental settings and the simulation environments used for highlighting the performance of our method. Sections \ref{sec:results} and \ref{sec:disc} describe the results on a number of simulation settings and also bring out the advantages as well as limitations of our approach. Section \ref{sec:cf} concludes the paper and points out future improvements for our method.


\section{The Method}\label{sec:method}
The objective of our work is to find a suitable policy (a sequence of actions given states of the environment) for UAV navigation that avoids obstacles (both stationary and mobile). We propose a general method which can find such suitable policies. Our method can be integrated with a high-level planner which is supplied with overall path objective, a start and a goal position. 

\subsection{Problem Definition}
In order to safely navigate without colliding against obstacles in an indoor or outdoor environment, the UAV needs to be aware of the state $s$ of the environment. 
The state of the environment is a tuple of properties of the environment which characterize it and aid the UAV in navigation. Once the state $s$ is known, the UAV selects an appropriate action $a$. The action the UAV chooses affects the visual information available to the UAV. In the obstacle avoidance problem, this means the UAV chooses to move in some particular direction leading to a change in its position, orientation and visual feedback. The UAV gets to observe more obstacles or perhaps more free space in front depending on this change in position and/or orientation. As noted in Section \ref{sec:intro}, the UAV needs to choose an action depending on the state at every instant $t$ when it navigates through the environment. Further, each action taken affects future states and hence future decisions of the UAV. Based on the action taken, the realization of the next state is probabilistic implying that navigation by avoiding obstacles is a sequential decision making problem in the face of uncertainty.

Prior works \cite{cad2rl,arxivref2} assume that the monocular image of the environment is a good indicator of the state of the environment. However, since the UAV's monocular camera has a limited field of vision, we believe that the UAV controller cannot infer the full state of the environment solely based on the RGB image. Instead, the UAV controller only has an estimate of the state and this estimate is formally known as an \emph{observation}. In the method we propose, the input to our model is a monocular RGB image without depth or other sensory inputs, whereas the \emph{observation} is the predicted depth map obtained from the monocular image. Based on these assumptions, we model the UAV obstacle avoidance and navigation problem in the framework of partially observable Markov decision processes (POMDPs).

We propose a POMDP model $\langle S,A,P,R,\Omega,\mathcal{O},\gamma \rangle$ for the obstacle avoidance problem. Here $S$ is the set of states of the environment, referred to as ``state space", while $A$ is the set of \emph{feasible} actions and referred to as the ``action space''. $P$ 
is the transition probability function that models the evolution of states based on actions chosen and is defined as $P: S\times A \times S\rightarrow [0,1]$. $R$ is the \emph{reinforcement} or the \emph{reward} function defined as $R: S \times A \rightarrow \R$. The reward function serves as a \emph{feedback} signal to the UAV for the action chosen. For instance, in a state $s$, if the UAV selects an action $a$ which steers it away from an obstacle, the reward for that state-action pair $(s,a)$ is positive, implying that the action $a$ is beneficial in the state $s$, while picking an action which results in collision will naturally yield a negative reward. $\Omega$ is the set of observations and an observation $o \in \Omega$ is an estimate of the true state $s$. $\mathcal{O}:S \times A \times \Omega \rightarrow [0,1]$ is a conditional probability distribution over $\Omega$, while $\gamma \in (0,1)$ is the discount factor. At each time $t$, the environment state is $s_t \in S$. The UAV takes an action $a_t \in A$ which causes the environment to transition to state $s_{t+1}$ with probability $P(s_{t+1}|s_t,a_t)$. Based on this transition, the UAV receives an observation $o_t \in \Omega$ which depends on $s_{t+1}$ with probability $\mathcal{O}(o_t|s_{t+1},a_t)$. The aim is to solve the obstacle avoidance problem, which translates to the task of finding an optimal \emph{policy} $\pi^*:\Omega \rightarrow A $. By determining an optimal policy, the UAV controller is able to select an action at each time step $t$ that maximizes the expected sum of discounted rewards, which is denoted as $\E \left[ \sum\limits_{t=0}^{\infty} \gamma^t R(s_t,a_t)\right]$.

\subsection{Model}
We need to define the sets $\Omega$, $A$ and the functions $R,P,\mathcal{O}$ in order to find an optimal policy. 
The input to our model is the monocular RGB image, without any depth information. Our model extracts the depth map from the RGB image which acts as the observation for the UAV controller. The depth map predicted from the RGB image indicates the distance between the objects and the UAV. Given an observation, the feasible actions ($A$) available for the UAV are ``go straight'', ``turn right'' and ``turn left''. The reward function $R$ is designed using the depth information and its exact analytical form is explained in Section \ref{sec:exp}. In order to determine the functions $P$ and $\mathcal{O}$, we must be aware of the structure of the environment and the motion dynamics of the UAV. In practice, these are impossible to know. , but the UAV must be capable of navigating in unknown, unstructured environments in the presence of other factors like wind, turbulence etc. Thus, we propose a Reinforcement learning technique to find an optimal policy for UAV navigation. Reinforcement learning is a model-free learning-based approach to solve (PO)MDPs when the model information via $P, R$ (and $\mathcal{O}$) is not available.

When model information is unavailable, one of the well known approaches learns an optimal policy using Q-values. The Q-value $Q^\pi(s,a)$ corresponding to the policy $\pi$ is defined as the expected sum of discounted rewards obtained by taking the action $a$ upon state $s$ and following the policy $\pi$ thereafter. The optimal Q-values are defined as $Q^*(s,a) = \max\limits_{\pi} Q^\pi(s,a)$. Once the optimum Q-values in a state $s$ are obtained, the optimal action is picked by finding $arg \max\limits_a Q(s,a)$. So, the optimal policy can be computed by finding the optimal action for every state. Q-learning \cite{ql}, is a model-free iterative algorithm to learn the optimal Q-value of every state-action $(s,a)$ pair. The Q-value update of any such pair is given below:
\begin{equation}
Q(s,a):=Q(s,a)+\alpha(r+\gamma \max_{a'}Q(s',a')-Q(s,a)).
\end{equation}
However, this algorithm suffers from \emph{curse of dimensionality}. This is because iterative learning the Q-values for huge state-space requires maintaining and updating Q-values for all unique state-action pairs which turns out to be computationally in-feasible. Deep Q-Networks (DQN) \cite{dqn} solve this issue by utilizing a neural network parametrized by weights ($w$) to approximate the Q-value (denoted as $Q(s,a|w)$) for a given state input. Experience replay improves the stability of the algorithm in which experience tuples $(s, a, r, s')$ are stored in a replay memory ($D$). During training, mini-batches of the experience are sampled uniformly and input to the network to calculate the Bellman residual as the loss given by 
\begin{equation}\label{eqn:lossfunc}
\begin{split}
L_i(w_i) = \E_{(s,a,r,s')\sim D}[ (r + \gamma \max\limits_{a'} Q(s',a';w^-_i) \\- Q(s,a;w_i) )^2 ].
\end{split}
\end{equation}
Here, $w^-$ represents weights of the target network which is an older copy of network weights lagging behind a few iterations. To achieve a better approximation, the weights are updated using mini-batch gradient descent.

Since an observation received in a POMDP is only the representative of the underlying environment state, $Q(o,a|w)\neq Q(s,a|w)$ holds. However, augmenting a recurrency to DQN integrates the observations over time to better estimate the underlying state, thereby narrowing the gap between $Q(o,a|w)$ and $Q(s,a|w)$ \cite{drqn}. Hence, we present a memory augmented convolutional neural network architecture to approximate the Q-values from the observations. The performance of the proposed architecture for UAV obstacle avoidance is analyzed in Section \ref{sec:results}.

\subsection{Deep Recurrent Q-Network with Temporal Attention}
The architecture for approximating Q-values is based on deep recurrent Q-network with attention. This approach essentially keeps track of the past few observations. In the UAV obstacle avoidance application, we keep track of the depth maps obtained from the RGB images. The recurrent network possesses the ability to learn temporal dependencies by using information from an arbitrarily long sequence of observations, while the temporal attention weighs each of the recent observations based on their importance in decision-making. 

At time $t$, the proposed model utilizes a sequence of recent $L$ observations ${o_{t-(L-1)},...,o_t}$. Each observation $o_{t-(L-i)}, \; 0 \leq i \leq L$ is a depth map which is processed by convolutional layers of the network, followed by a fully connected layer augmented with LSTM \cite{lstm} recurrent network layer. The DRQN model with LSTM estimates the Q-value $Q(o_t,h_{t-1},a_t)$, where $h_{t-1}$ is the hidden state of the recurrent network and is determined 
 as $h_{t-1}=LSTM(h_{t-2},o_{t-1})$. The hidden state represents the information gathered over time. 
 
 \begin{figure}[t]
\begin{center}
\includegraphics[scale=0.4]{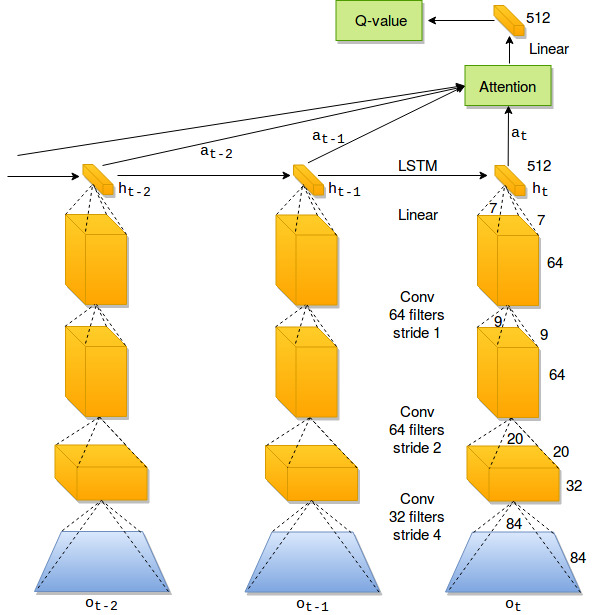}
\caption{\small Control Network: Architecture of Deep Recurrent Q-Network with Temporal Attention. Number of filters, stride and output size are mentioned for each convolutional layer.}
\label{fig:darqn}
\end{center}
\end{figure}
 
Following the LSTM layer, we propose the use of Temporal Attention \cite{temporal} in our model for evaluating the informativeness of each observation in the sequence. Temporal attention optimizes a weight vector with values depicting the importance of observations at the previous instants. This increases the training speed and provides better generalizability over the training dataset. Let $(\mathbf{v}_{t-(L-1)},\mathbf{v}_{t-(L-2)},\ldots,\mathbf{v}_t)^\top$ be the vector of feature vectors obtained from the convolutional layers, over the past $L$ observations. For each $ 1 \leq i \leq L $, ${v}_{t-(L-i)}$ is a feature vector in $\R^{m \times 1}$. The vector of weights $(e_{t-(L-1)},e_{t-(L-2)},\ldots,e_{t})^\top$, for the $L$ feature vectors, is computed using the obtained hidden state values and the feature vector given by: 
\begin{equation}
\label{eqn:tanh}
e_{t-(L-i)} = \mathbf{w}^\top\tanh \left(W_ah_{t-1} + U_a \mathbf{v}_{t-(L-i)} + b_a\right)
\end{equation}
in which $\mathbf{w},b_a \in \R^{a\times 1},W_a \in \R^{a\times r},U_a \in \R^{a \times m}$ are all learnable parameters and $h_t \in \R^{r \times 1}$. In \eqref{eqn:tanh}, $\tanh(\cdot)$ is an activation function which is computed for every element of the vector given by $W_ah_{t-1} + U_a \mathbf{v}_{t-(L-i)} + b_a$. Here, we assume that $r$ is the size of an RNN hidden state, $m$ is the encoding size of CNN and $a$ is the attention matrix size. The $\tanh$ activation function is applied pointwise on the vector obtained from $W_ah_{t-1} + U_a \mathbf{v}_{t-(L-i)} + b_a$.

These weights are normalized using the softmax function
\begin{equation}
a_{t-(L-i)} = \frac{\exp(e_{t-(L-i)})}{\sum\limits_{j=1}^{L} \exp(e_{t-(L-j)})}.
\end{equation}
Further, to predict the Q-values a context vector is computed using the above calculated softmaxes and hidden states as:
$$\phi(t) = \sum\limits_{j=1}^{L} (a_{t-(L-j)} v_{t-(L-j)}). $$
The obtained context vector is input to a single fully connected layer with ReLU\cite{relu} activation functions that outputs approximated Q-value for each action. The complete model is trained by minimizing a loss function as described in \cite{drqn}. The proposed model using temporal attention is illustrated in Fig \ref{fig:darqn}.

\subsection{Obtaining depth maps from RGB images}
\begin{figure}[h!]
\begin{center}
\includegraphics[scale=0.27]{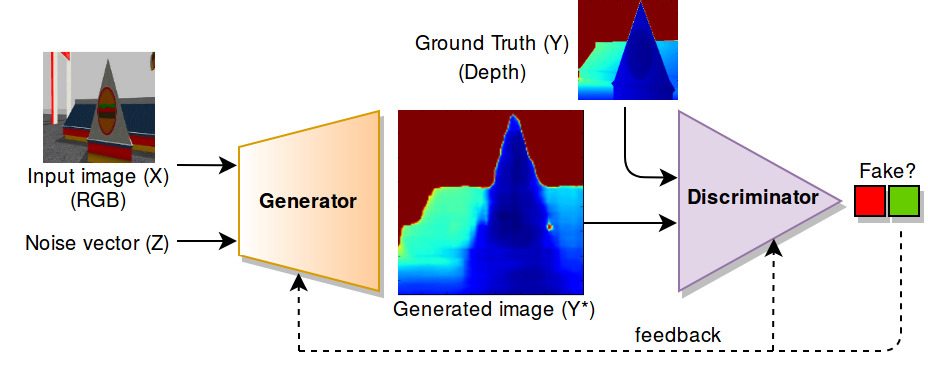}
  \caption{\small Depth Network: Conditional GAN architecture}
  \label{fig:pix2pix}
\end{center}
\end{figure}

The UAV's on-board sensor is limited to providing monocular RGB image data. Effective depth prediction from an RGB image is essential when operating in the physical world. Learning a mapping for image translation $X\rightarrow Y$, given image pairs $\{x \in X,y \in Y\}$, is a challenging task in the computer vision community. In this work, we propose the use of conditional generative adversarial network (cGAN) \cite{pix2pix} for this image-to-image translation. This approach uses two separate ConvNets (called as Generator and Discriminator) with BatchNorm layers and ReLU activation layers. The Generator (G) ConvNet is an encoder-decoder structure with skip connections, designed to generate realistic fake images taking $x\in X$ and a noise vector z as inputs. The Discriminator (D) network classifies randomly picked images as fake or real with a cross-entropy loss. Let $\theta_D$ and $\theta_G$ represent the weights of the Discriminator and Generator networks respectively. The Generator is expected to produce images close to the ground truth, while the discriminator is supposed to distinguish between fake images and the real images. Hence in a sense, the objectives of these two networks are opposed to each other. The loss function $\mathcal{L}_{cGAN}(\theta_G,\theta_D)$ defined below reflects these objectives:
\begin{equation}
\begin{split}
\label{discloss}
\mathcal{L}_{cGAN}(\theta_G,\theta_D) = \mathbb{E}_{x,y \sim p_{data}} [\log D(x,y)] + \\
\mathbb{E}_{x\sim p_{data}(x), z \sim p_z(z)} [\log(1-D(x,G(x,z)))].
\end{split}
\end{equation}
In the above equation, the variable $x$ is the RGB image and $y$ is its true depth map. The depth map generated by G is denoted as $G(x,z)$. $D(x,y)$ and $D(x,G(x,z))$ are the probabilities of the image belonging to the real class.
Training a cGAN involves a few steps. Initially, the discriminator is trained on real and fake depth images with the correct labels for few epochs. Following this, the generator is trained using the real/fake predictions from the trained discriminator as its objective. This procedure is repeated for few epochs until the generated fake depth maps are difficult to distinguish from the real depth maps. The cGAN architecture is illustrated in Fig.\ref{fig:pix2pix}. The approach also incorporates $L_1$ loss to generate better near ground truth images.
\begin{equation}
\begin{split}
\label{l-one}
\mathcal{L}_{L_1}(\theta_G) = \mathbb{E}_{x,y \sim p_{data}, z \sim p_z(z)} [\parallel y-G(x,z) \parallel_{1}].
\end{split}
\end{equation}
Hence the final objective of the model can be analytically represented as
\begin{equation}
\label{gan_loss}
\min_{\theta_G} \max_{\theta_D} \{\mathcal{L}_{cGAN} (\theta_G,\theta_D) + \lambda \mathcal{L}_{L_1}(\theta_G)\},
\end{equation}
where $\lambda$ is an adjustable hyper-parameter. In contrast to previous methods (\cite{fcrn}-\cite{semisupdeeplearningdepth}) our approach learns a loss function adaptable to the input data, making it domain independent and suitable for our problem of intermediate depth prediction for obstacle avoidance.

\subsection{Remarks}
\begin{enumerate}
 \item 
It must be noted that the depth maps generated from cGANs as described above still provide limited information with respect to the visual geometry of the environment surrounding the UAV (a similar problem when monocular camera images are used). This issue of partial information was highlighted in Section \ref{sec:intro}. The limited information obtained in stages from cGAN can be stored and collected to make a better navigation decision. The task of using all the relevant partial information obtained in the past is done by the LSTM network architecture as described earlier in this Section.
\item The deep RL method we propose in this section learns optimal Q-values and the optimal policy for the obstacle avoidance task. There are also other policy improvement approaches for learning a good policy. Recently proposed methods like the Asynchronous Advantage Actor-Critic (A3C) \cite{a3c}, deep deterministic policy gradient (DDPG) \cite{ddpg} and dueling network architecture for double deep Q-networks (D3QN) \cite{d3qn}  can also be used with our proposed method. For using these methods, one has to change the loss function \eqref{eqn:lossfunc} for the network architecture. 
Our method involving temporal attention can be easily integrated with A3C, DDPG and D3QN. However, in this paper, our objective is to highlight the need for using LSTM architecture for partially observable scenarios in UAV obstacle avoidance.
\end{enumerate}

\section{Experimental Setup}\label{sec:exp}
\subsection{Depth Network Settings}
\label{subsec:dns}
\begin{figure}[t!]
\begin{center}
\includegraphics[scale=0.06]{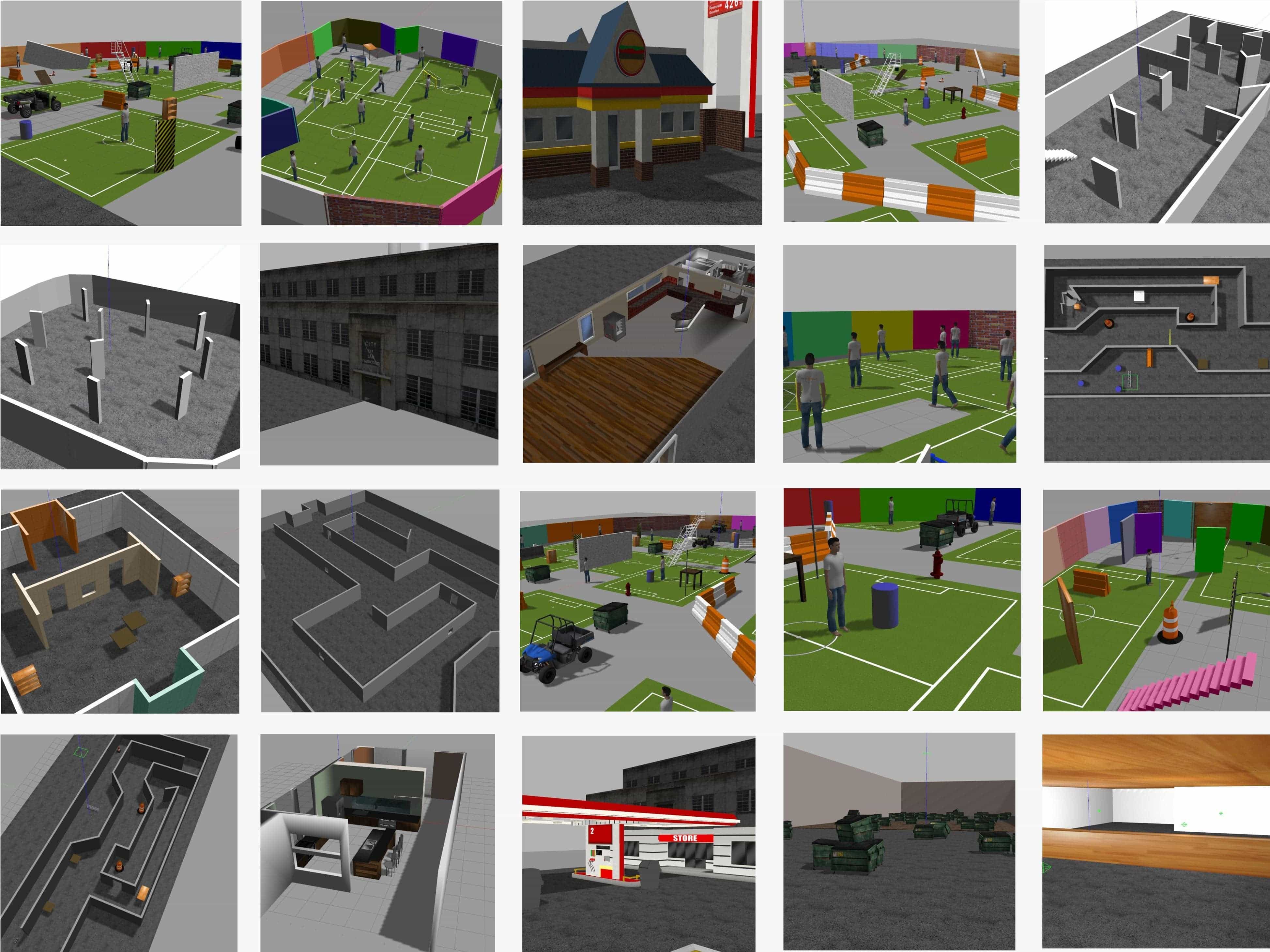}
  \caption{\small Screenshots of the designed environments in Gazebo. We cover a large range of colors, textures, sizes and shapes for obstacles and walls.}
  \label{fig:collage}
\end{center}
\end{figure}

The proposed conditional GAN is initially trained on a total of $90,000$ RGB-D image pairs collected from the Gazebo\cite{gazebo} simulated environments each having different characteristics. We have a total of 22 different simulated indoor environments, of which few are inspired from \cite{punarjay} while the rest are self designed. The environments consist of broad and narrow hallways, small and large enclosed areas with floorings ranging from asphalt to artificial turf. The simulated environments also contain structured and unstructured obstacles like humans, traffic cones, tables etc., placed at random positions and with random orientation. The walls and obstacles with diverse shapes, textures and colours provide abundant visual information for effective learning.  Fig. \ref{fig:collage} shows example snapshots of the environment.
 
 The RGB-D image pairs are collected using a Kinect sensor mounted on the flying drone in simulation, covering all possible viewpoints. Further, the dataset is augmented off-line by random flipping, adding random jitter and random alteration to the brightness, saturation, contrast and sharpness. The network is trained on the entire collected dataset for $20$ epochs in batches of size $4$ on an NVIDIA Titan X machine. 
 We require the depth network (trained on the simulated images) to predict depth from the unseen real-world images. Predicting depth from either simulated images or real-world images are similar tasks. Thus, it is intuitive to leverage the low-level features learned during training in one task for a different, yet similar task. The basic idea in fine-tuning of depth architecture is exactly this. Once a neural network has been trained on simulated images, the lower layers of the neural network are frozen (so that features learned are kept intact). Then, using the real images, one can just retrain the output layer. By freezing the lower layers, we are using the same features learned earlier to predict depth on the real-world images. The major benefit of this approach is that the network works effectively on similar tasks without the need for training from scratch and also requires substantially low data. In our problem, the network is fine-tuned using $8,000$ and $16,000$ augmented pairs from RGBD-human-explore data \cite{ram-lab} and NYU2 dataset \cite{nyu2}, respectively.
 
\subsection{Control Network and Simulation Settings}
For RL algorithms to learn an effective collision avoidance policy, the UAV learning agent must have enough experience of undesirable events like collision. Training a learning algorithm on a fragile drone in a physical environment is expensive and hence the performance of DRL algorithms is usually demonstrated on simulated environments. In this work, we build and test our UAV collision avoidance algorithms on the aforementioned simulated environments.
Our method initially trains the UAV by starting off with simple hallway environments free of obstacles. Gradually the environment complexity is increased by narrowing down the pathways, enclosing the free space and increasing the density of obstacles. The proposed control network is trained to learn the observation-action value over the last $L$ observations (depth images received from the simulated Kinect sensor aboard the UAV) corresponding to the three actions \emph{``go straight"}, \emph{``turn left"} and \emph{``turn right"}, respectively.  The agent receives a reward after each step and the reward function is defined as
$$R_i=\min\bigg(1,\frac{d_i - r_{drone}}{\sigma - r_{drone}}\bigg),$$ 
where $d_i$ is the distance to the nearest obstacle at the $i^{th}$ decision making instant, $r_{drone}$ is the radius of the drone which is set to $0.292$m and $\sigma$ is the threshold distance which is set to $1.5$m. The reward function shown above penalizes the action of the controller when it is at a distance less than $\sigma-r_{drone}$ from the obstacle. If the agent collides, the episode ends with a penalty of $-10$. Otherwise, the episode continues until it reaches the maximum number(1000) of steps and terminates with no penalty. The agent also receives an additional $+0.5$ reward if it chooses the ``go straight" action.  The bias for the ``go straight" action helps the UAV to always move forward and turn only when there are obstacles in its clear view.
Additionally, to cope with the exploration-exploitation tradeoff, a linear annealed policy is utilized during training with initially chosen value of $\epsilon$ = 1 that drops eventually to 0.05 as the final value. The network hyper-parameter values are as shown in Table \ref{hyperparam}. 
\begin{table}[h]
\begin{center}
\begin{tabular}{ c | c }
  Entity & Value  \\
  \hline
  Discount Factor ($\gamma$) & 0.99  \\
  Mini-batch size & 32  \\
  Learning rate & 0.0001  \\
  Target network update frequency & 400 \\
  Input observation size & 84$\times$84 \\
  Conv1 layer filter size & 8$\times$8 \\
  Conv2 layer filter size & 4$\times$4 \\
  Conv3 layer filter size & 3$\times$3 \\
\end{tabular}
\caption{\small Hyper-parameter values of proposed control network}
\label{hyperparam}
\end{center}
\end{table}

For the proposed control network to be applicable for robotic applications, the learned policy should be effectively transferable to the real-physical systems. However, this is highly challenging because of the huge gap in visual information available in the real and simulated worlds. Moreover, the depth maps produced by the proposed depth network are too noisy when compared to depth images obtained from the simulated kinect sensor. To overcome this, we degrade the sensor images with Gaussian blurring, random jitter and superpixel replace (replacement probability 0.5) at the time of training. This additional noise is crucial for non-linear function approximators like neural networks to learn and generalize well, making them robust and transferable to real-world systems.

\section{Experimental Results}\label{sec:results}



\subsection{Depth network performance on monocular RGB images}
The depth network is trained as mentioned in the previous section. Once trained, we evaluate the performance of the depth network for two measures - the inference speed and the depth prediction quality, respectively. The inference rate of deep learning models is critical when applied to robotic applications, especially when solving for effective collision avoidance models in flying robots. We tested our model on an NVIDIA GeForce GTX 1050 mobile GPU with 8 GB RAM and Intel core i7 processor machine and observed a sufficient enough data rate of 20Hz on an average. In addition, we also implemented previously used depth network in robotic applications \cite{fcrn} and noted an inference rate of 1.4Hz on the same machine configuration. 

To assess depth prediction quality of the cGAN architecture, we evaluate the network on unseen simulated data and the fine-tuned data (real-world images) ($5,000$ and $2,500$ samples respectively). For evaluation, we compute $L_1$ and cGAN loss which has been demonstrated to be a better loss function to generate near ground truth images \cite{pix2pix}. Table \ref{loss} depicts the network performance in various scenarios. 
\begin{table*}[t!]
\begin{center}
  \begin{tabular}{|c|c|c|c|c|c|}
  \hline
      \multicolumn{2}{|c|}{Scenario} &
      \multicolumn{2}{c|}{Training loss} &
      \multicolumn{2}{c|}{Testing loss} \\
    \hline
      Training set & Testing set & $L_1$ & cGAN & $L_1$ & cGAN\\
      \hline
     simulated & simulated & 0.106 & 0.666 & 1.114 & 0.711 \\
    simulated (same training as previous case) & real-world &0.106 & 0.666 & 2.779 & 0.738 \\
     simulated + real-world & real-world & 0.135 & 0.692 & 1.792 & 0.695 \\
    \hline
  \end{tabular}
  \end{center}
  \caption{\small Depth network's quantitative analysis}
  \label{loss}
\end{table*}

\begin{figure}[t!]
\begin{center}
\includegraphics[scale=0.12]{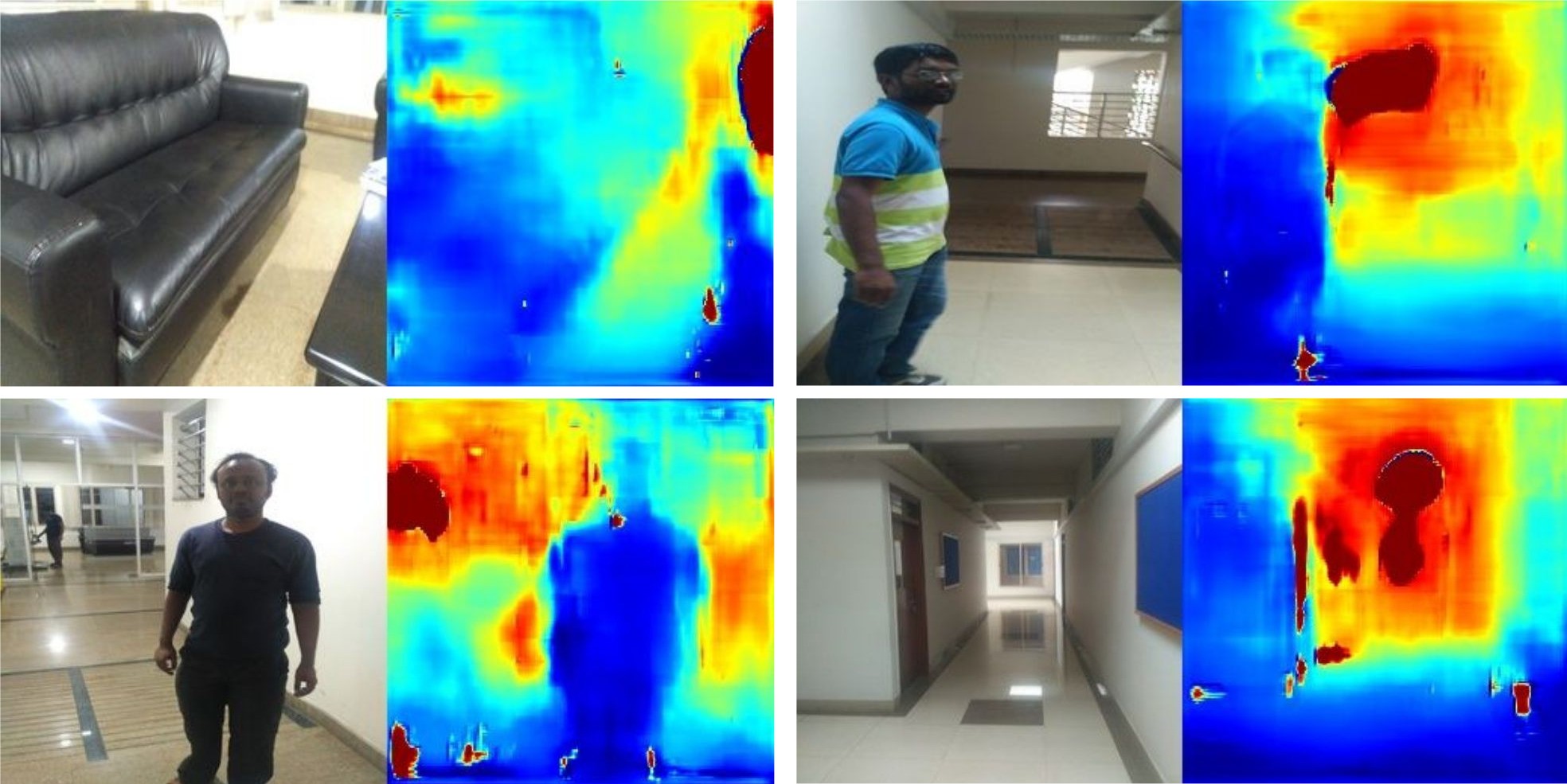}
  \caption{\small Example of depth maps generated by the proposed network (trained on simulated data) for completely unseen real world data with variable illumination, color and texture (Red: far, Blue: near) }
  \label{realenv}
\end{center}
\end{figure}

The first row of values depicts the training and testing loss on manually collected data (data collection is explained in Section \ref{subsec:dns}).
The second row depicts training loss on our simulated dataset, while the testing loss is on a mix of images from the NYU2 \cite{nyu2} and  RGBD-human-explore \cite{ram-lab} datasets. The third row of values corresponds to the case where the network was trained entirely on the simulated data with fine-tuning. The results in the third row show that such a trained network possesses the ability to generalize well on real-world data. Fig. \ref{realenv} showcases some samples of the depth maps generated by the cGAN network. The sample images are taken at the Department of Computer Science and Automation, Indian Institute of Science (IISc) and consist of humans (imitating obstacles) and hallways with varying illumination, colour and texture which the network has never seen before. The quantitative and qualitative evaluation depicts that the proposed model provides a remarkable boost to the data cycle rate which is essential in robotic applications and can be effectively transferred to real-world systems.

\subsection{Control network evaluation}
\begin{figure*}[h]
\begin{subfigure}[b]{0.5\textwidth}
\centering
\includegraphics[scale=0.48]{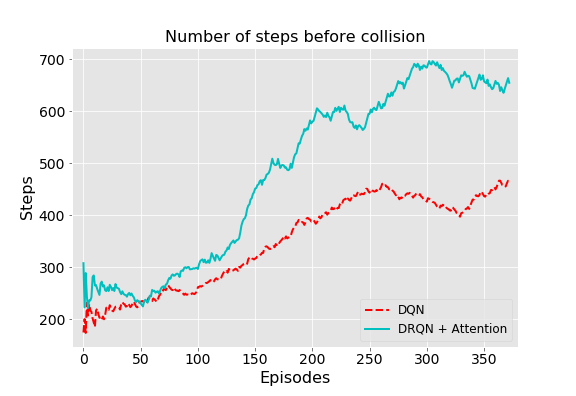}
  \caption*{(a)}
  \label{fig:tr1}
\end{subfigure}%
\begin{subfigure}[b]{0.5\textwidth}
\centering
\includegraphics[scale=0.48]{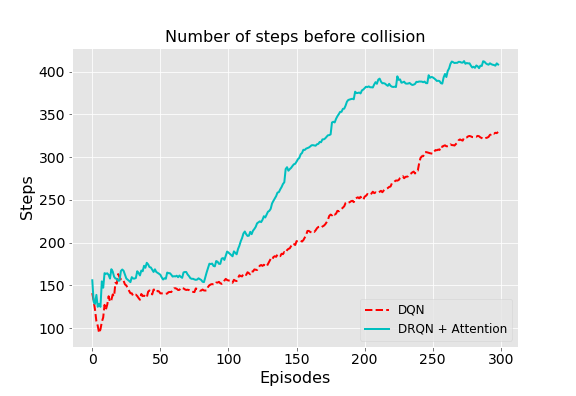}
  \caption*{(b)}
  \label{fig:tr2}
\end{subfigure}
\caption{Training learning curve of the proposed network and DQN for two different environment settings: (a) An open area with scattered static obstacles of varying sizes and structures (b) Maze like environment with narrow pathways and no scattered obstacles}
\label{fig:dqn-drqna}
\end{figure*}

\begin{figure}[h]
\begin{center}
\includegraphics[scale=0.34]{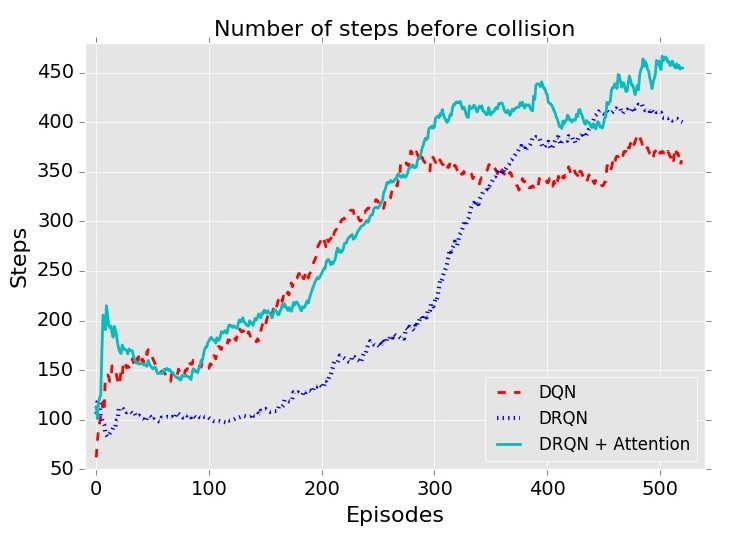}
  \caption{\small Training learning curve of the proposed network and DQN for an environment consisting of an enclosed area with scattered static obstacles of varying sizes and structures}
  \label{fig:dqn-drqna-darqn}
\end{center}
\end{figure}

We evaluate the performance of the proposed control network i.e., Deep Recurrent Q-network with Temporal Attention, and compare it with the baseline DQN previously proposed \cite{cad2rl}. We also implement two other policies - random and straight. The random policy picks an action with equal probability for each observation, while the straight policy always picks the ``go straight" action. The metric used for performance evaluation is the average number of steps taken until collision with an obstacle. Both the DQN and our proposed model are trained in 12 different simulated indoor environments comprising of hallways and rooms with obstacles of varying structures and sizes. Some snapshots of these environments were illustrated in the earlier section.  Figures \ref{fig:dqn-drqna} and\ref{fig:dqn-drqna-darqn} show the learning curves during training for both the algorithms for three different environments. These graphs depict the number of steps the UAV takes until collision. Fig. \ref{fig:dqn-drqna-darqn} also shows the performance of DRQN for one such environment.

As can be observed, partial observability of the environment hinders the performance of DQN in the obstacle avoidance problem. However, the graph shows that augmenting a memory network with attention is beneficial as it retains crucial information gathered over time and this gives an additional fillip to the learning when compared to the no-attention counterpart.

\subsubsection{Testing in Simulated environments}
The trained models are tested on six randomly selected simulated environments out of the twelve environments used for training. The network takes the noisy depth map and outputs the UAV control signal. The output control signal is expected to safely navigate the UAV within the environment for longer duration. Out of the six environments used for testing, three comprise of enclosed areas with randomly scattered static obstacles of varying sizes and structures (named as Env-1, Env-2 and Env-3 in Table \ref{result_tab}). 
The fourth environment (Env-4) is a maze like structure with narrow pathways and no scattered obstacles. The fifth environment (Env-5) is a small enclosed area having poles in between. The sixth environment (Env-6) simulates a cafe-like environment and has 7 human actors randomly walking inside the cafe. The actors are not programmed to avoid the moving UAV and their movement paths are completely random. For this cafe-like environment, the model is initially trained with 3 human actors (randomly moving, not designed to avoid the UAV), but tested with 7 moving actors. We analyze the model performance for 200 episodes in each environment and Table \ref{result_tab} indicates the average number of steps the UAV takes until collision as well as the standard error. From Table \ref{result_tab}, it can be seen that using our approach, the UAV flies for the maximum number of time instants until collision. 
\subsubsection{Results}
\begin{figure*}[h!]
\begin{center}
\includegraphics[scale=0.6]{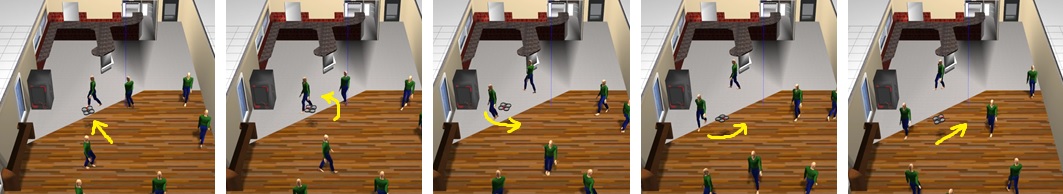}
  \caption{\small Snapshots of UAV avoiding randomly moving human actors. The yellow arrows show the path the UAV selects in order to avoid the obstacle.}
  \label{snapshot}
\end{center}
\end{figure*}
A snapshot of the testing setup is demonstrated in Fig. \ref{snapshot}, depicting the learned UAV model maneuvering in Env-6, effectively avoiding the randomly moving human actors inside a cafe.
The proposed DRL model also observes a notable inference rate of 60 Hz on NVIDIA GeForce GTX 1050 mobile GPU, essential for robotic applications.

\begin{figure*}[h!]
\begin{center}
\includegraphics[scale=0.7]{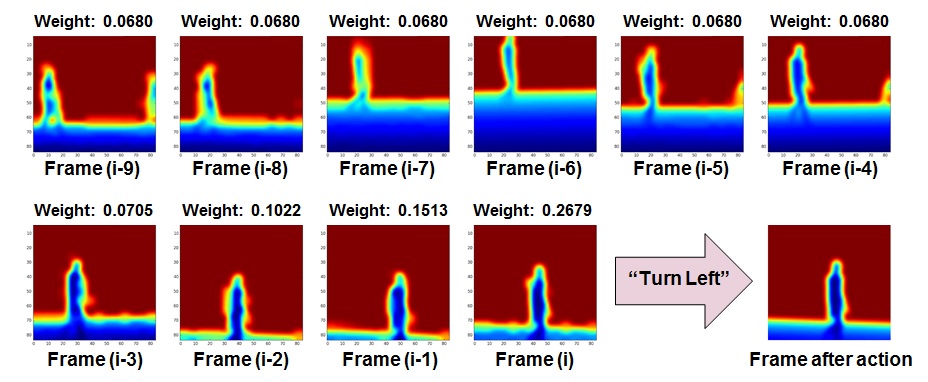}
  \caption{\small Temporal Attention weights over the most recent $L$=10 observations}
  \label{taweights}
\end{center}
\end{figure*}

Fig. \ref{taweights} illustrates the weights attributed to a sequence of $10$ images over the recent past used to find the UAV's next move. It can be analyzed from the images that in an environment consisting of non-stationary obstacles, predicting the direction of the next step based only on the recent observation (for instance Frame (i) in Fig. \ref{taweights}) is a complicated task. Possessing a memory facilitates an agent to infer the direction of the moving obstacle (such as a human actor walking right) and thereby performing an appropriate action (``turn left") to avoid collision. It is important to note that our proposed algorithm outperforms DQN on different environments. The advantages of the policy learnt by our method are: (i) the UAV smoothly follows a path while avoiding static obstacles and (ii) in the presence of dynamic obstacles which obstruct the UAV's view, the UAV skillfully chooses actions to avoid collisions with the dynamic obstacles as well. Video results from these experiments can be seen at \url{https://bit.ly/2PNgWsk}

A UAV is a power-constrained system. Thus, a navigation and obstacle avoidance method must be designed in such a manner that it uses the available battery power judiciously. 
We say that a UAV wobbles when it takes a long sequence of consecutive left and right turns which do not lead to displacement in its position. Thus, the UAV does not cover any distance when it wobbles, but still, power is consumed in this sequence of right-left movements. This motion without displacement is minimized by our method, which naturally leads to a reduction in power wastage.
In order to test for energy efficiency, we designed a simulation environment and tested the proposed method as well as the previously proposed algorithm D3QN \cite{arxivref2} over it. The simulated environment consists of straight hallway with two $45\degree$ turns in between. The navigation task considered is episodic, wherein the UAV starts at a pre-specified initial position. An achievable destination point after the second turn is also specified and the episode terminates when the UAV reaches this destination point. Based on the drop in the battery level and the distance covered, we compute the energy consumption per meter values for both methods by using the power rating of the battery. We observed that for this simulated environment, the average energy consumption over several runs is $0.0571$ Wh/m for our approach and $0.0743$ Wh/m for D3QN. Thus, this shows that our method achieves a lower value of energy consumption per unit distance traveled when compared to the D3QN method. 

\begin{table*}[h!]
\centering
\begin{tabular}{|c|c|c|c|c|c|c|}
\hline
 & Env-1 & Env-2 & Env-3 & Env-4 & Env-5 & Env-6 \\ \hline
Straight & 61$\pm$16 & 58$\pm$14 &  76$\pm$23 & 65$\pm$12 & 42$\pm$12 & 27$\pm$9 \\
Random & 125$\pm$84 & 176$\pm$121 &  113$\pm$83 & 162$\pm$88 & 121$\pm$76 & 42$\pm$19 \\
DQN & 207$\pm$103 & 229$\pm$95 &  286$\pm$142 & 634$\pm$241 & 384$\pm$126 & 162$\pm$83 \\
D3QN & 248$\pm$109 & 271$\pm$104 &  297$\pm$133 & 658$\pm$253 & 414$\pm$146 & 177$\pm$85 \\
Our Approach & 323$\pm$134 & 342$\pm$131 &  326$\pm$156 & 764$\pm$273 & 652$\pm$243 & 247$\pm$77 \\ \hline
\end{tabular}
\caption{Results indicating the average number of steps taken by UAV (along with standard deviation) until collision}
\label{result_tab}

\end{table*}

\section{Discussion}
\label{sec:disc}

Our method has multiple advantages as well as some limitations that we list below:
\begin{itemize}
\item Our approach of adopting cGAN architecture for depth prediction in autonomous aerial systems is novel. Notably, the proposed approach is trained entirely on simulated data and with little fine-tuning on the NYU2 and RGBD-human-explore dataset. The results validate that the model is highly generalizable and qualifies to be adopted in real world applications. As demonstrated by our results, the remarkably high inference rate and transferability of the approach makes it a suitable candidate for intelligent robotic applications.

\item We show in our experiments that augmenting DRL with memory networks and temporal attention facilitates the agent to retain vital information gathered from the past observations. This aids the agent towards making better and informed decisions. This learning ability benefits the autonomous agent to maneuver safely in environments without prior knowledge of the surroundings, as well as in environments with moving obstacles. Furthermore, the agent is competent to move deftly near corners (refer supplementary video) which has been found to be a challenging task for the previously proposed controllers (\cite{punarjay}, \cite{cad2rl}).

\item The reward function is designed by considering the energy constraints on aerial systems and time factor in navigation tasks. The bias towards the ``go straight'' action in the reward function ensures that the UAV maintains its course except when avoiding obstacles in its field of view. In addition, when compared to the D3QN approach, the proposed controller gives smoother trajectories and UAV wobbling is minimized that would otherwise cause a lot of energy to be wasted which is highly undesirable in UAV applications. Our control method minimizes this power wastage and yields considerable power savings. The bias towards the ``go straight'' action might be problematic at intersections, where the UAV has to turn right or left. However, we would like to emphasize that our proposed method handles only obstacle avoidance and can be easily integrated with a high-level path planner that handles the computation of the path from start to goal position.

\item Although the proposed depth prediction network learnt to predict depth maps from the unseen physical world images, the results are noisy. The control network trained with the manually-added noise generalizes and adapts to the noise. However, there is scope for improvement as far as the depth network is concerned. Training depth network on visually high-fidelity simulated data can yield smoother depth predictions.
\end{itemize}

\section{Conclusions and Future Work}
\label{sec:cf}

In this paper, we design and analyze the performance of a Deep Recurrent Q-Network  with Temporal Attention which is utilized by a deep RL robotic controller for effective obstacle avoidance of UAV in cluttered and unseen environments. The proposed method first utilizes the cGAN network to predict the depth map from a monocular RGB image which is then used to decide the optimal action. The method addresses the problem of partial observability in obstacle avoidance by retaining the crucial information over the long sequence of observations. Experimental results over various settings exhibit significant improvements over Deep Q-Network(DQN) and D3QN algorithms. A potential future direction for our work would be to improve the visual quality of images generated by the cGAN architecture. In GAN architectures, the discriminator block captures the class-specific content from images without imposing constraints on the visual quality of the generated images. The cGAN architecture can be made to generate good quality images by suitably modifying the loss function. Some similarity indices which guarantee structural integrity (e.g., multiscale structural-similarity MS-SIM) can be used for this purpose (see \cite{future1}). Another future enhancement would be to use different GAN architectures for depth prediction (see \cite{future2,survey1}). 

The proposed obstacle avoidance method is seen to work well in avoiding obstacles in indoor environments (see Section \ref{sec:results}). However, we would also like to test the performance in real outdoor environments. A possible exciting line of research can be to learn concise abstractions of history in recurrent networks, sufficient for optimal decision-making. We would also like to incorporate scene prediction \cite{future3} to learn better navigation controls for avoiding obstacles. Regret minimization is another criterion used in RL. Though it has been explored for games like VizDoom and Minecraft \cite{rdrl}, the same has not been explored in robotics. It will be interesting to see what policies guarantee low regret in UAV obstacle avoidance and how such policies can be interpreted.

\bibliographystyle{IEEEtran}
\bibliography{root}

\begin{thebibliography}{10}
\providecommand{\url}[1]{#1}
\csname url@samestyle\endcsname
\providecommand{\newblock}{\relax}
\providecommand{\bibinfo}[2]{#2}
\providecommand{\BIBentrySTDinterwordspacing}{\spaceskip=0pt\relax}
\providecommand{\BIBentryALTinterwordstretchfactor}{4}
\providecommand{\BIBentryALTinterwordspacing}{\spaceskip=\fontdimen2\font plus
\BIBentryALTinterwordstretchfactor\fontdimen3\font minus
  \fontdimen4\font\relax}
\providecommand{\BIBforeignlanguage}[2]{{%
\expandafter\ifx\csname l@#1\endcsname\relax
\typeout{** WARNING: IEEEtran.bst: No hyphenation pattern has been}%
\typeout{** loaded for the language `#1'. Using the pattern for}%
\typeout{** the default language instead.}%
\else
\language=\csname l@#1\endcsname
\fi
#2}}
\providecommand{\BIBdecl}{\relax}
\BIBdecl

\bibitem{crop}
M.~Reinecke and T.~Prinsloo, ``The influence of drone monitoring on crop health
  and harvest size,'' in \emph{2017 1st International Conference on Next
  Generation Computing Applications (NextComp)}, July 2017, pp. 5--10.

\bibitem{infra}
Y.~Ham, K.~K. Han, J.~J. Lin, and M.~Golparvar-Fard, ``Visual monitoring of
  civil infrastructure systems via camera-equipped {U}nmanned {A}erial
  {V}ehicles {UAV}s: {A} review of related works,'' \emph{Visualization in
  Engineering}, vol.~4, no.~1, p.~1, Jan 2016.

\bibitem{disaster}
M.~Erdelj, E.~Natalizio, K.~R. Chowdhury, and I.~F. Akyildiz, ``{H}elp from the
  sky: leveraging {UAV}s for disaster management,'' \emph{IEEE Pervasive
  Computing}, vol.~16, no.~1, pp. 24--32, Jan 2017.

\bibitem{delivery}
\BIBentryALTinterwordspacing
P.~Grippa, D.~A. Behrens, C.~Bettstetter, and F.~Wall, ``{J}ob selection in a
  network of autonomous {UAV}s for delivery of goods,'' \emph{CoRR}, vol.
  abs/1604.04180, 2016. [Online]. Available:
  \url{http://arxiv.org/abs/1604.04180}
\BIBentrySTDinterwordspacing

\bibitem{filming}
\BIBentryALTinterwordspacing
M.~Funaki and N.~Hirasawa, ``Outline of a small unmanned aerial vehicle
  (ant-plane) designed for antarctic research,'' \emph{Polar Science}, vol.~2,
  no.~2, pp. 129 -- 142, 2008. [Online]. Available:
  \url{http://www.sciencedirect.com/science/article/pii/S1873965208000236}
\BIBentrySTDinterwordspacing

\bibitem{kinect}
Z.~Zhang, ``Microsoft {K}inect sensor and its effect,'' \emph{IEEE MultiMedia},
  vol.~19, no.~2, pp. 4--10, Feb 2012.

\bibitem{slam-rgbd}
W.~G. Aguilar, G.~A. Rodr{\'i}guez, L.~{\'A}lvarez, S.~Sandoval, F.~Quisaguano,
  and A.~Limaico, ``Visual slam with a {RGB-D} camera on a quadrotor {UAV}
  using on-board processing,'' in \emph{Advances in Computational
  Intelligence}, I.~Rojas, G.~Joya, and A.~Catala, Eds.\hskip 1em plus 0.5em
  minus 0.4em\relax Cham: Springer International Publishing, 2017, pp.
  596--606.

\bibitem{slam-lidar-stereo}
T.~Gee, J.~James, W.~V.~D. Mark, P.~Delmas, and G.~Gimel'farb, ``Lidar guided
  stereo simultaneous localization and mapping ({SLAM}) for {UAV} outdoor 3-{D}
  scene reconstruction,'' in \emph{2016 International Conference on Image and
  Vision Computing New Zealand (IVCNZ)}, Nov 2016, pp. 1--6.

\bibitem{sfm-opt-flow}
\BIBentryALTinterwordspacing
D.-J. Lee, P.~Merrell, Z.~Wei, and B.~E. Nelson, ``Two-frame structure from
  motion using optical flow probability distributions for unmanned air vehicle
  obstacle avoidance,'' \emph{Machine Vision and Applications}, vol.~21, no.~3,
  pp. 229--240, Apr 2010. [Online]. Available:
  \url{https://doi.org/10.1007/s00138-008-0148-9}
\BIBentrySTDinterwordspacing

\bibitem{sfm-monocular}
H.~Alvarez, L.~M. Paz, J.~Sturm, and D.~Cremers, ``Collision avoidance for
  quadrotors with a monocular camera,'' in \emph{Experimental Robotics}.\hskip
  1em plus 0.5em minus 0.4em\relax Springer, 2016, pp. 195--209.

\bibitem{realtimeslam}
J.~Li, Y.~Bi, M.~Lan, H.~Qin, M.~Shan, F.~Lin, and B.~M. Chen, ``Real-time
  simultaneous localization and mapping for {UAV}: A survey,'' in
  \emph{International Micro Air Vehicle Conference and Competition (IMAV)},
  2010.

\bibitem{only-kinect}
N.~Eric and J.~W. Jang, ``Kinect depth sensor for computer vision applications
  in autonomous vehicles,'' in \emph{2017 Ninth International Conference on
  Ubiquitous and Future Networks (ICUFN)}, July 2017, pp. 531--535.

\bibitem{only-opt-flow}
S.~Zingg, D.~Scaramuzza, S.~Weiss, and R.~Siegwart, ``{MAV} navigation through
  indoor corridors using optical flow,'' in \emph{2010 IEEE International
  Conference on Robotics and Automation}, May 2010, pp. 3361--3368.

\bibitem{fcrn}
I.~Laina, C.~Rupprecht, V.~Belagiannis, F.~Tombari, and N.~Navab, ``{D}eeper
  depth prediction with fully convolutional residual networks,'' in \emph{2016
  Fourth International Conference on 3D Vision (3DV)}, Oct 2016, pp. 239--248.

\bibitem{fast-monocular}
M.~Mancini, G.~Costante, P.~Valigi, and T.~A. Ciarfuglia, ``{F}ast robust
  monocular depth estimation for obstacle detection with fully convolutional
  networks,'' in \emph{2016 IEEE/RSJ International Conference on Intelligent
  Robots and Systems (IROS)}, Oct 2016, pp. 4296--4303.

\bibitem{field-monocular}
F.~Liu, C.~Shen, G.~Lin, and I.~Reid, ``{L}earning depth from single monocular
  images using deep convolutional neural fields,'' \emph{IEEE Transactions on
  Pattern Analysis and Machine Intelligence}, vol.~38, no.~10, pp. 2024--2039,
  Oct 2016.

\bibitem{semisupdeeplearningdepth}
Y.~Kuznietsov, J.~St{\"u}ckler, and B.~Leibe, ``{S}emi-supervised deep learning
  for monocular depth map prediction,'' in \emph{Proc. of the IEEE Conference
  on Computer Vision and Pattern Recognition}, 2017, pp. 6647--6655.

\bibitem{egomotion}
T.~Zhou, M.~Brown, N.~Snavely, and D.~G. Lowe, ``Unsupervised learning of depth
  and ego-motion from video,'' in \emph{2017 IEEE Conference on Computer Vision
  and Pattern Recognition (CVPR)}, July 2017, pp. 6612--6619.

\bibitem{punarjay}
P.~Chakravarty, K.~Kelchtermans, T.~Roussel, S.~Wellens, T.~Tuytelaars, and
  L.~V. Eycken, ``{CNN}-based single image obstacle avoidance on a quadrotor,''
  in \emph{2017 IEEE International Conference on Robotics and Automation
  (ICRA)}, May 2017, pp. 6369--6374.

\bibitem{jmod}
M.~Mancini, G.~Costante, P.~Valigi, and T.~A. Ciarfuglia, ``{J-MOD}2: {J}oint
  monocular obstacle detection and depth estimation,'' \emph{IEEE Robotics and
  Automation Letters}, vol.~3, no.~3, pp. 1490--1497, July 2018.

\bibitem{intermediate}
\BIBentryALTinterwordspacing
S.~Yang, S.~Konam, C.~Ma, S.~Rosenthal, M.~M. Veloso, and S.~Scherer,
  ``{O}bstacle avoidance through deep networks based intermediate perception,''
  \emph{CoRR}, vol. abs/1704.08759, 2017. [Online]. Available:
  \url{http://arxiv.org/abs/1704.08759}
\BIBentrySTDinterwordspacing

\bibitem{crash-learn}
D.~Gandhi, L.~Pinto, and A.~Gupta, ``Learning to fly by crashing,'' in
  \emph{2017 IEEE/RSJ International Conference on Intelligent Robots and
  Systems (IROS)}, Sept 2017, pp. 3948--3955.

\bibitem{cad2rl}
\BIBentryALTinterwordspacing
F.~Sadeghi and S.~Levine, ``{CAD2RL:} {R}eal single-image flight without a
  single real image,'' in \emph{Robotics: Science and Systems XIII,
  Massachusetts Institute of Technology, Cambridge, Massachusetts, USA, July
  12-16, 2017}, 2017. [Online]. Available:
  \url{http://www.roboticsproceedings.org/rss13/p34.html}
\BIBentrySTDinterwordspacing

\bibitem{arxivref2}
L.~Xie, S.~Wang, A.~Markham, and N.~Trigoni, ``Towards monocular vision based
  obstacle avoidance through deep reinforcement learning,'' in \emph{RSS 2017
  workshop on New Frontiers for Deep Learning in Robotics}, 2017.

\bibitem{fps}
\BIBentryALTinterwordspacing
G.~Lample and D.~S. Chaplot, ``Playing {FPS} games with deep reinforcement
  learning,'' 2017. [Online]. Available:
  \url{https://aaai.org/ocs/index.php/AAAI/AAAI17/paper/view/14456}
\BIBentrySTDinterwordspacing

\bibitem{BertB}
D.~Bertsekas, \emph{Dynamic Programming and Optimal Control},
  $4^{th}$~ed.\hskip 1em plus 0.5em minus 0.4em\relax Belmont,MA: Athena
  Scientific, 2013, vol.~II.

\bibitem{ql}
C.~J. Watkins and P.~Dayan, ``Q-learning,'' \emph{Machine learning}, vol.~8,
  no. 3-4, pp. 279--292, 1992.

\bibitem{dqn}
V.~Mnih, K.~Kavukcuoglu, D.~Silver, A.~A. Rusu, J.~Veness, M.~G. Bellemare,
  A.~Graves, M.~Riedmiller, A.~K. Fidjeland, G.~Ostrovski \emph{et~al.},
  ``Human-level control through deep reinforcement learning,'' \emph{Nature},
  vol. 518, no. 7540, p. 529, 2015.

\bibitem{drqn}
M.~Hausknecht and P.~Stone, ``Deep recurrent {Q}-learning for partially
  observable {MDPs},'' in \emph{2015 AAAI Fall Symposium Series}, 2015.

\bibitem{temporal}
W.~Pei, T.~Baltrušaitis, D.~M.~J. Tax, and L.~P. Morency, ``Temporal
  attention-gated model for robust sequence classification,'' in \emph{2017
  IEEE Conference on Computer Vision and Pattern Recognition (CVPR)}, July
  2017, pp. 820--829.

\bibitem{lstm}
S.~Hochreiter and J.~Schmidhuber, ``Long short-term memory,'' \emph{Neural
  computation}, vol.~9, no.~8, pp. 1735--1780, 1997.

\bibitem{nyu2}
\BIBentryALTinterwordspacing
N.~Silberman, D.~Hoiem, P.~Kohli, and R.~Fergus, ``Indoor segmentation and
  support inference from {RGBD} images,'' in \emph{Proceedings of the 12th
  European Conference on Computer Vision - Volume Part V}, ser. ECCV'12.\hskip
  1em plus 0.5em minus 0.4em\relax Berlin, Heidelberg: Springer-Verlag, 2012,
  pp. 746--760. [Online]. Available:
  \url{http://dx.doi.org/10.1007/978-3-642-33715-4_54}
\BIBentrySTDinterwordspacing

\bibitem{ram-lab}
\BIBentryALTinterwordspacing
L.~Tai, S.~Li, and M.~Liu, ``A deep-network solution towards model-less
  obstacle avoidance,'' in \emph{2016 IEEE/RSJ International Conference on
  Intelligent Robots and Systems (IROS)}, Oct 2016, pp. 2759--2764. [Online].
  Available: \url{https://ram-lab.com/dataset/rgbd-human-explore.tar.gz}
\BIBentrySTDinterwordspacing

\bibitem{pix2pix}
P.~Isola, J.~Y. Zhu, T.~Zhou, and A.~A. Efros, ``{I}mage-to-image translation
  with conditional adversarial networks,'' in \emph{2017 IEEE Conference on
  Computer Vision and Pattern Recognition (CVPR)}, July 2017, pp. 5967--5976.

\bibitem{relu}
\BIBentryALTinterwordspacing
V.~Nair and G.~E. Hinton, ``{R}ectified linear units improve restricted
  {B}oltzmann machines,'' in \emph{Proceedings of the 27th International
  Conference on International Conference on Machine Learning}, ser.
  ICML'10.\hskip 1em plus 0.5em minus 0.4em\relax USA: Omnipress, 2010, pp.
  807--814. [Online]. Available:
  \url{http://dl.acm.org/citation.cfm?id=3104322.3104425}
\BIBentrySTDinterwordspacing

\bibitem{a3c}
V.~Mnih, A.~P. Badia, M.~Mirza, A.~Graves, T.~Lillicrap, T.~Harley, D.~Silver,
  and K.~Kavucuoglu, ``Asynchronous methods for deep reinforcement learning,''
  in \emph{Proceedings of the 33rd International Conference on International
  Conference on Machine Learning}, ser. Proceedings of Machine Learning
  Research, vol.~48.\hskip 1em plus 0.5em minus 0.4em\relax PMLR, 2016, pp.
  1928--1937.

\bibitem{ddpg}
\BIBentryALTinterwordspacing
T.~P. Lillicrap, J.~J. Hunt, A.~Pritzel, N.~Heess, T.~Erez, Y.~Tassa,
  D.~Silver, and D.~Wierstra, ``Continuous control with deep reinforcement
  learning,'' \emph{CoRR}, vol. abs/1509.02971, 2015. [Online]. Available:
  \url{http://arxiv.org/abs/1509.02971}
\BIBentrySTDinterwordspacing

\bibitem{d3qn}
H.~Van~Hasselt, A.~Guez, and D.~Silver, ``Deep reinforcement learning with
  double q-learning.'' in \emph{AAAI}, vol.~2.\hskip 1em plus 0.5em minus
  0.4em\relax Phoenix, AZ, 2016, p.~5.

\bibitem{gazebo}
``{Gazebo Simulator},'' \url{www.gazebosim.org}.

\bibitem{future1}
\BIBentryALTinterwordspacing
K.~Ridgeway, J.~Snell, B.~Roads, R.~S. Zemel, and M.~C. Mozer, ``Learning to
  generate images with perceptual similarity metrics,'' \emph{CoRR}, vol.
  abs/1511.06409, 2015. [Online]. Available:
  \url{http://arxiv.org/abs/1511.06409}
\BIBentrySTDinterwordspacing

\bibitem{future2}
\BIBentryALTinterwordspacing
D.~Berthelot, T.~Schumm, and L.~Metz, ``{BEGAN:} boundary equilibrium
  generative adversarial networks,'' \emph{CoRR}, vol. abs/1703.10717, 2017.
  [Online]. Available: \url{http://arxiv.org/abs/1703.10717}
\BIBentrySTDinterwordspacing

\bibitem{survey1}
\BIBentryALTinterwordspacing
A.~Creswell, T.~White, V.~Dumoulin, K.~Arulkumaran, B.~Sengupta, and A.~A.
  Bharath, ``Generative adversarial networks: An overview,'' \emph{CoRR}, vol.
  abs/1710.07035, 2017. [Online]. Available:
  \url{http://arxiv.org/abs/1710.07035}
\BIBentrySTDinterwordspacing

\bibitem{future3}
C.~Sheppard and M.~Rahnemoonfar, ``Real-time scene understanding for uav
  imagery based on deep convolutional neural networks,'' in \emph{2017 IEEE
  International Geoscience and Remote Sensing Symposium (IGARSS)}, July 2017,
  pp. 2243--2246.

\bibitem{rdrl}
\BIBentryALTinterwordspacing
P.~H. Jin, S.~Levine, and K.~Keutzer, ``Regret minimization for partially
  observable deep reinforcement learning,'' \emph{CoRR}, vol. abs/1710.11424,
  2017. [Online]. Available: \url{http://arxiv.org/abs/1710.11424}
\BIBentrySTDinterwordspacing

\end{thebibliography}

\end{document}